\documentclass[10pt]{article}

\usepackage{graphicx}
\usepackage{geometry}
\usepackage[utf8]{inputenc}
\usepackage{bussproofs}
\usepackage{mathpartir}
\usepackage{graphicx}
\usepackage{tikz-cd}
\usepackage{amsmath, amsthm, amssymb, amsfonts, bm}
\usepackage{booktabs}
\usepackage{adjustbox}
\usepackage{units}
\usepackage{lipsum}
\usepackage{algpseudocode}
\usepackage{tkz-euclide}

\usepackage{amsmath,amsfonts,bm}

\def\eqref#1{equation~\ref{#1}}

\def\1{\bm{1}}

\DeclareMathAlphabet{\mathsfit}{\encodingdefault}{\sfdefault}{m}{sl}
\SetMathAlphabet{\mathsfit}{bold}{\encodingdefault}{\sfdefault}{bx}{n}

\newcommand{\Var}{\mathrm{Var}}

\newcommand{\Ebb}{\mathbb{E}}

\newcommand{\Ac}{\mathcal{A}}
\newcommand{\Sc}{\mathcal{S}}

\theoremstyle{plain}
\newtheorem{theorem}{Theorem}[section]
\newtheorem{lemma}[theorem]{Lemma}

\theoremstyle{definition}

\numberwithin{figure}{section}
\numberwithin{equation}{section}

\newcommand{\ol}{\overline}

\usepackage[utf8]{inputenc} 
\usepackage[T1]{fontenc}    
\usepackage{hyperref}
\hypersetup{colorlinks,allcolors=black}
\usepackage{url}            
\usepackage{booktabs}       
\usepackage{amsfonts}       
\usepackage{nicefrac}       
\usepackage{microtype}      
\usepackage{xcolor}         
\usepackage{natbib}

\title{Second order dynamic mean field programming}

\author{%
  George Stamatescu\\
  School of Computer Science\\
  University of Adelaide\\
  Adelaide, SA 5001 \\
  \texttt{george.stamatescu@adelaide.edu.au} \\
}

\title{Dynamic Mean Field Programming}

\author{
  George Stamatescu\\
}

\begin{document}

\maketitle

\vskip 0.15in

\begin{abstract}%
A dynamic mean field theory is developed for finite state and action Bayesian reinforcement learning in the large state space limit. In an analogy with statistical physics, the Bellman equation is studied as a disordered dynamical system; the Markov decision process transition probabilities are interpreted as couplings and the value functions as deterministic spins that evolve dynamically. Thus, the mean-rewards and transition probabilities are considered to be quenched random variables. The theory reveals that, under certain assumptions, the state-action values are statistically independent across state-action pairs in the asymptotic state space limit, and provides the form of the distribution exactly. The results hold in the finite and discounted infinite horizon settings, for both value iteration and policy evaluation. The state-action value statistics can be computed from a set of mean field equations, which we call dynamic mean field programming (DMFP). For policy evaluation the equations are exact. For value iteration, approximate equations are obtained by appealing to extreme value theory or bounds. The result provides analytic insight into the statistical structure of tabular reinforcement learning, for example revealing the conditions under which reinforcement learning is equivalent to a set of independent multi-armed bandit problems. 
\end{abstract}

\section{Introduction}
An agent interacting with a Markov decision process (MDP) with unknown parameters faces the problem of both estimating these parameters and determining a control strategy. A fundamental question concerns the efficient use of samples to balance the two goals, a question with a long history \cite{duff2002optimal}. In the case that there is only one state, the problem is known as the multi-armed bandit, for which there exist a variety of solution concepts \cite{jacko2019finite}, based on Bayesian or frequentist estimation under finite or infinite horizon objectives. 

Index policies are a popular form of solutions to the bandit problem, where at each time step an index function is calculated for each arm (or action), and the arm with largest index is chosen. Index functions, such as the Gittins index in the Bayes-optimal approach or upper confidence bounds in the asymptotically optimal approach \cite{chang1987optimal} \cite{kaufmann2012bayesian}  \cite{russo2021note}, are functions of the empirical estimates or statistics of each arm's mean reward. Index policies, under a criteria of asymptotic optimality, can be thought of as principled heuristics \cite{duff2002optimal}, and are a motivation for the current paper.

In a general MDP with multiple states, one would also like to form an index policy over the state-action or Q-values. Since obtaining samples of the optimal value functions directly is not generally possible, one approach is to attempt to translate the estimates and uncertainty of the MDP parameters, which can be sampled, to the state-action value functions via the Bellman equation. This paper revisits the question of how uncertainty translates or propagates through the Bellman equation in the Bayesian setting, borrowing ideas and tools from statistical physics to study this question.

Under the Bayesian perspective the Q-values are random variables, since they can be written as functions of the MDP parameters. It is generally thought that since the Q-values depend on the parameters and each other recursively via the Bellman equation, this makes exact calculation of the posterior over Q-values analytically impossible. Here we show that the problem simplifies in the large state space limit, provided the prior or posterior distributions over MDP parameters satisfies certain conditions.

Our starting point for the theory is the observation that the Bellman equation can be seen as analogous to the dynamics of a spin glass system or equivalently, the dynamics of a recurrent neural network in the forward evaluation \cite{Castellani_2005},  \cite{poole2016exponential}, \cite{helias2019statistical}. In the language of statistical physics, we interpret value iteration for an MDP, with parameters drawn from a prior or posterior, as a deterministic dynamical system subject to \emph{quenched disorder} \cite{crisanti2018path}. The disorder in this context arises from the uncertainty over transition probabilities and mean-rewards. In the spin glass analogy, the transition probabilities represent couplings and the Q-value functions represent the local fields for deterministic spins. In the neural network language, these are connection weights and activation inputs for neurons, respectively.

Inspired by this analogy we develop a dynamic mean field theory (DMFT) for the Bellman equation, in the sense of \cite{sompolinsky1988chaos}. This theory was originally derived for random recurrent neural networks in the continuous time setting, which has recently been revisited and extended to modern deep neural networks \cite{segadlo2022unified}. While the key steps in the theory and its results hold for the Bellman equation, there are significant differences between the systems, making the derivation presented here novel.

The basis of DMFT is the generating functional or path integral formalism for disordered systems. This approach studies a moment generating function (MGF) for a suitably defined probability distribution over the Q-value iterates (or plain Q-values in the finite horizon case). This MGF is itself random due to the uncertainty in the transition probabilities and mean-rewards. We carefully detail the statistical assumptions necessary for the DMFT to hold in Section \ref{sec:bayesian_RL}, however the two key assumptions we highlight are that (i) the prior or posterior over MDP parameters factorises over state-action pairs, and (ii) that the transitions have, what amounts to, a flat Dirichlet distribution over the simplex. These assumptions are commonly made in reinforcement learning theory and will both hold at the start of learning. The former remains true throughout learning, while the latter breaks down, as most MDPs have sparse state transition matrices. We discuss at length the assumptions, their breakdown, and the consequences for the mean field theory and what corrections might be made based on more advanced techniques from field theory.

The assumptions underpin a large deviations principle, which we conjecture to be true, for the probability distribution over the Q-value iterates in the infinite horizon case (or Q-values in the finite horizon case). In statistical physics terminology, this principle is equivalent to assuming that the MGF is a \emph{self-averaging} quantity, meaning that it concentrates in measure on its average \cite{touchette_harris}. If we take this conjecture to be true, one can calculate statistics for a typical realisation of the MDP parameters from the averaged MGF. We are able to perform this average due to an old result in statistics from \cite{vonNeumann1941}, which allows us to write the generating function expansion for a linear combination of Dirichlet random variables, also known as a finite Dirichlet mean \cite{pitman2018random} \cite{hjort2005exact}.

Once the MGF has been averaged, we are able to establish the asymptotic independence of Q-values, in both finite and discounted infinite horizon, via the application of the method of steepest descents, or saddlepoint approximation. This method is applied to a transformation of the averaged moment generating function via suitably defined ``auxilliary fields", which produces an integral which can be asymptotically expanded \cite{bender1999advanced}.

The equations for the saddlepoint produce the so-called mean field equations, which in the MDP context we name the dynamic mean field programming equations. The computation of these equations provides the exact statistics, such as the mean and variance, of the Q-values. In our flat Dirichlet setting, we are simply propagating a mean through the equations, whose expression is very similar to the dynamic programming equations.

From the mean field statistics, an index policy can be derived in principle, although as discussed the flat Dirichlet assumption will hold only at the start of learning in general. Even so, the result already provides practical insight: at the start of learning, an agent holding these prior beliefs faces a set of $N$ independent multi-armed bandit problems, in each state. The duration of this mean field effect will of course break down during learning, for example if the MDP has sparse transition matrices. The practical consequence is that either the prior assumptions should be changed, to incorporate sparsity, or that the algorithm deployed under such assumptions should be a plain multi-armed bandit algorithm in each state, acting locally and not planning ahead.

In order to compute the DMFP equations explicitly we derive the approximate equations using extreme value theory, whose accuracy improves with increasing action space size. In the i.i.d case, which corresponds to having identical priors across transition probabilities and mean-rewards, we find closed form expressions for the mean field equations. In the general case, during learning, we resort to numerical approximations based on extreme value theory. We provide simulation evidence in support of the results, confirming the mean field theory and the accuracy of the approximate mean field equations.

Since the language and techniques of statistical physics would be foreign to most reinforcement learning researchers, we have strived to clarify the language used and include a healthy number of references. The dynamic mean field theory is essentially a method involving only a few key steps, which we hope to have carefully presented and discussed at length. Fortunately, in comparison the spin glass case from which it originates, we enjoy some simplifications in our setting, since the Bellman equation evolves on a discrete index (either iterates or time steps) and we thus do not have to deal with the technical details of a continuous time system, though some new complications are introduced in our setting as well. 



\subsection{Outline}

The paper proceeds as follows. We introduce our notation for Bayesian reinforcement learning in Section \ref{sec:bayesian_RL}. In Section \ref{sec:analogy} we present the basic analogy between random neural networks and Bayesian reinforcement learning. We also provide a brief overview of mean field theory as encountered in machine learning, placing dynamic mean field theory in the broader gamut. In Section \ref{sec:DMFT_BayesRL} we develop the dynamic mean field theory for RL, which provides a systematic derivation of the DMFP equations. In Section \ref{sec:DMFP} we present the dynamic mean field programming (DMFP) equations and derive approximate computations based on extreme value theory, bounds and heuristics. Throughout the paper we focus on the discounted infinite horizon case, since the finite horizon and other settings follow the same form. We discuss the results in Section \ref{sec:discussion}.

\section{Bayesian reinforcement learning}\label{sec:bayesian_RL}

Let us introduce our notation for Markov decision processes (MDPs) and  Bayesian reinforcement learning. An infinite horizon, discounted MDP $M = (\mathcal{S}, \mathcal{A}, P, r, \beta)$, is specified by a state space $\mathcal{S}$, an action space $\mathcal{A}$; a transition function $P : \mathcal{S} \times \mathcal{A}\to \Delta(\mathcal{S})$, where $\Delta(\mathcal{S})$ the probability simplex and we say $P_{s'| sa}$ is the probability of transitioning into state $s'$ upon taking action a in state $s$; a random reward function $r : \mathcal{S} \times \mathcal{A} \to\mathbb{R}$, denoted $r_{sa}$ with mean $\rho_{sa}$ and finite variance, and a discount factor $\beta \in [0, 1)$, which defines an effective horizon for the problem. We denote the size of the state space as $|\mathcal{S}| = N$

A policy specifies a decision-making strategy. We restrict attention to the set of stationary policies $\pi : \mathcal{S} \to \Delta(\mathcal{A})$. Given a fixed policy and fixed starting state $s_0 = s$, we define the state-action or Q-value function $Q^\pi : \mathcal{S} \times \mathcal{A} \to \mathbb{R}$ as the discounted sum of future rewards
\begin{align}
    Q^\pi_{sa} = \mathbb{E}[\sum_{t=1}^\infty \beta^t r_{s_t a_t} | \pi, s_0 = s, a_0 = a ]
\end{align}
where expectation is with respect to the randomness of the trajectory, that is, the randomness in state transitions and $\pi$. The solution to the Bellman optimality equation are referred to as the optimal Q-values. In the case the parameters are known, we may use dynamic or linear programming methods to compute solutions. In the case that the MDP parameters are unknown an agent must balance exploration and exploitation. We adopt a Bayesian approach to estimation, and make the following assumptions.

{\bf Assumption 1. }(Mean-reward and transition independence assumption.) We assume independence across the mean-rewards and transition functions $\rho_{sa} \perp \rho_{(sa)'}$, $P_{sa} \perp P_{(sa)'}$ and $\rho_{sa} \perp P_{(sa)'} $ for all pairs $sa$ and $(sa)'$.

{\bf Assumption 2. }(Flat Dirichlet)
We assume the transition probabilities are given by a flat Dirichlet random vectors for each state-action pair, $P_{sa} \sim \text{Dirichlet}(\alpha_{sa})$, meaning that the parameters are of the order $\frac{\alpha_{s'|sa} }{\sum_{s''} \alpha_{s''|sa}} \sim \mathcal{O}(1/N)$, for all $s'$ and $(sa)$ pairs.

{\bf Assumption 3. }(Arbitrary mean-reward distribution)
We assume the mean-reward $\rho_{sa}$ has an arbitrary distribution with a moment generating function that exists. We denote its mean and variance as $\mu_{\rho_{sa}}$ and $\sigma^2_{\rho_{sa}}$. Notably, we do not assume bounded rewards.

{\bf Assumption 4. }(Bounded effective horizon)
We assume $\beta \in [0,1)$ is fixed, meaning that the effective horizon $H_\text{eff} = \frac{1}{1-\beta}$ is bounded.

The flat Dirchlet assumption implies that each Q-value statistically depends on all other Q-values directly, with physicists using the term interaction or coupling. This is crucial for the mean field effects to arise. It is in general a strong assumption for reinforcement learning that would not apply to all problems except at the start of learning under a flat prior. It is possible that this assumption may be weakened to introduce a degree of sparsity but this question is left for future work.  The implication of the fifth assumption, of bounded horizon (or bounded effective horizon) is that we can assume it does not scale with the number of states $N$.

\subsection{Related work in reinforcement learning}



A line of research in both the frequentist and Bayesian literature has analysed exploration bonuses for sample efficient RL. In the Bayesian case the exploration bonus can be derived from an upper bound or approximation to the posterior variance of the Q-values, for example \cite{dearden1998bayesian}, \cite{sorg}, \cite{odonoghue18uncertainty}. Similarly, \cite{OsbandVanRoyPSRLvsOptimism} use a bound on the posterior variance to heuristically propose a Gaussian approximation for posterior sampling. We anticipate the mean-field setting developed here to be a useful test case for these algorithms.

Recent work has applied an epistemic risk-seeking approach to develop Bayesian reinforcement learning with regret guarantees in finite horizon problems  \cite{O'DonoghueVariational} \cite{O'Donoghue2020Making}. The approach essentially bounds an object related to the cumulant generating function of the posterior over Q-values, which physicists would recognise as the free energy. The bounds are restricted to the finite horizon case, or the case where the MDP is a directed acyclic graph. In contrast, under our assumptions, the mean field result is exact in the large system limit and applies to infinite horizon problems. Our theory may be relevant to risk sensitive approaches since it determines the cumulants for Q-values exactly. We should note that physicists have long studied the interplay between field theoretic, perturbative and variational approaches \cite{opper2001naive} \cite{bachschmid2016variational}. 

Similar to this Bayesian work are frequentist approaches to derive exploration bonuses \cite{neuPikeBurke}, which often have some relationship or resemblance to the empirical variance of the Q-values. In the discussion we point to possible connections by switching from the Bayesian perspective to considering about frequentist estimators. It was this line of thinking that motivated the initial investigations of this paper.

%



\section{An analogy between reinforcement learning and neural networks} \label{sec:analogy}

The analogy we work from is most clear when presenting side by side the equations for Q-value iteration for the discounted infinite horizon problem and random recurrent neural networks. Denote $x_i(t)$ as neurons at time $t$, $\theta_i$ as thresholds, weights (or couplings) $J_{ij}$ are here assumed to be Gaussian and i.i.d and we denote $\phi(\cdot)$ as a non-linearity, for example $\tanh(\cdot)$. The equations are,
\begin{align}
   &Q_{sa}^{k+1} = \rho_{sa} + \beta \sum_{s'} P_{s'|s,a} \max_{a'} Q_{s'a'}^{k}\label{eq:QvalIteration}\\
 &x_i(t+1) = \theta_i + \sum_j J_{ij} \phi(x_j(t)) \label{eq:randomRNN}
\end{align} 
 By inspection, state-action pairs $(sa)$ correspond to the site indices $i$ in the neural network case. Thus each transition probability couples a Q-value to a set of other Q-values. The non-linearity $\phi(\cdot)$ is typically a smooth function or at least piece-wise linear, where as the maximum function is neither. Couplings for neural networks are often taken as i.i.d and Gaussian, whereas the transition probabilities are drawn from Dirichlet distribution, and exhibit negative dependence. In order to obtain interesting limits, the mean and variance of the $J_{ij}$ are set to scale as $\bar{J}/N$ and $\sigma_J^2/N$ respectively. In the MDP case, the couplings are naturally scaled as they sum to one. 

 The analogy with neural networks is generally more helpful for the present context rather than the spin glass case, since the discrete time steps or layers correspond to iterations or time steps of dynamic programming, and the distribution of ``connection weights" is adapted during learning. More specifically, spin glass theory considers the distributions over couplings between spins to be fixed. For example independent and identically distributed (i.i.d) Gaussian couplings give the models of \cite{sompolinsky1982relaxational}, \cite{sompolinsky1988chaos}, and the symmetric versions, where the couplings to and from spin $i$ to spin $j$ are equal, correspond to the famous Sherrington-Kirkpatrick model \cite{mezard1987spin}. In the deep neural network case, the connection weights are asymmetric and typically drawn i.i.d from a Gaussian prior at the start of learning\footnote{Feedforward architectures also have independent weights across layers, whereas the recurrent architectures the weights are shared, resembling more closely the Bellman equation in the discounted infinite horizon setting.} but then proceed to evolve under stochastic gradient descent \cite{gerbelot2022rigorous}, with dependence emerging between weights. 

Beliefs over transition probabilities in the Bayesian reinforcement learning literature are usually described using Dirichlet distributions, as we do so here. The chief reason is for this choice is to exploit the multinomial-Dirichlet conjugacy for Bayesian updating \cite{duff2002optimal} \cite{poupart2006analytic}. As an agent interacts with a system, their posterior beliefs will update in the parts of the state-action space that they explore, while the independence assumptions of the prior are preserved. In contrast, under neural network learning, all weights are adapted simultaneously and the initial independence of the prior washes away.

The prior independence assumption in reinforcement learning, Assumption 1 above, is common in theoretical, tabular reinforcement learning and analogous to the i.i.d connection weights and thresholds for neural networks or spin glasses. The independence over state-action pairs is crucial to the dynamic mean field theory, and underlies the large deviations principle for the Bellman equation which we conjecture here.

The flat Dirichlet assumption, in this paper Assumption 2, is analogous to the fully or densely connected assumption of neural networks and mean-field spin glasses. This assumption is usually satisfied at the start of learning, but will only remain true throughout learning for a certain class of MDP, where one can transition to any other state. Such an MDP is essentially a type of contextual bandit, where the context is a discrete state index evolving not as a Markov chain but instead arriving i.i.d.  Although the flat Dirichlet assumption is a strong and restrictive one, we insist that the theory developed here should not be thought of as only fit for describing the start of reinforcement learning, or just a theory for a peculiar class of MDP. Rather, we view the present work as a first stepping stone toward a more precise understanding of how statistical assumptions on the model parameters of the MDP translate to statistical assumptions on the state-action values throughout learning. Later in the paper we discuss how tools from physics and large deviations that may allow us to systematically go beyond the mean field setting developed here.


\subsection{Static and dynamic mean field theory}\label{sec:phys}

Mean field theory and statistical physics have been commonly encountered in machine learning within the context of variational Bayesian inference \cite{wainwright2008graphical}. The idea is that one attempts to approximate an intractable posterior distribution using a simpler one whose parameters may be computed using optimisation algorithms. This approach is borrowed from statistical physics, which can be recognised when interpreting the posterior of interest as a Boltzmann-Gibbs distribution for a system with a given Hamiltonian, an energy function which defines the true interactions between the variables of interest. The Boltzmann-Gibbs distribution describes a physical system in \emph{thermal equilibrium}. This is often referred to as a static theory, which is not to say that such systems have no dynamics, but that equilibrium is the long-time limit to which appropriately defined dynamics relax towards. Dynamic theories study systems out of equilibrium, which in general may never relax towards a steady state.

A large deviations principle underpins rigorous mean field theories  \cite{touchette2009}. The principle essentially allow one to conclude for a given system with interaction that whilst the variables of interest may be dependent on each other, they are weakly dependent enough that as the system gets large its elements are effectively independent. In the static case, we say that there exists an \emph{effective} Hamiltonian, with a simplified interaction ``field" between the variables, such as each variable only depending on the mean of other variables. 
Disordered systems, such as spin glasses, are complex varieties of systems where the interactions between the spins are themselves random but considered to be fixed on a timescale of interest. Physicists and probabilists refer to such systems as having quenched randomness. Both statics and dynamics have been studied for various disordered systems.


A new point of contact between machine learning and a dynamic theory of disordered systems recently occurred in a series of papers which has helped establish initialisation theory for deep learning \cite{poole2016exponential} \cite{schoenholzdeep}. Originally developed in a series of papers for spin glasses and neuroscience \cite{sompolinsky1982relaxational} \cite{sompolinsky1988chaos}, this theory applies generally to systems where the couplings between fluctuating or deterministic spins are random and asymmetric. The asymmetry implies that equilibrium is not guaranteed since Markov dynamics for the system do not satisfy detailed balance. 
Like static mean field theory, the idea in the dynamic case is to replace the local field for variables in the dynamical system with an effective field with simplified interactions. The results show that neurons are effectively subject to a Gaussian field, whose statistics are given by a self-consistent, time-dependent mean field equations. See \cite{moynot2002large} for an accessible review of large deviations theory corresponding exactly to the dynamic mean field theory of recurrent neural networks in discrete time. We now proceed to develop the same approach in the Bayesian reinforcement learning setting, first outlining the tools (or tricks) physicists use in to derive the theory.

\section{A toolkit from statistical physics and probability}

In order to make the derivation in the next section more accessible to those without a background in statistical physics of disordered systems, we provide here a brief review of the key tools we use. Also included is the result of von Neumann on linear combinations of flat Dirichlet variables \cite{vonNeumann1941}, derived using the so-called Beta-Gamma algebra \cite{pitman2018random}, itself an old result with a long history of rediscovery \cite{ConniffeSpencer}.

\subsection{Fourier representation of the Dirac delta and the Hubbard-Stratonovich transformation}

A common tool used throughout physics and this paper is the Fourier integral representation of the Dirac delta $\delta(x)$,
\begin{align}\label{eq:fourier_dirac}
    \delta(x)  = \frac{1}{2\pi } \int_{-\infty}^{\infty} d\tilde{x} \exp(i\tilde{x}x)
\end{align}
We will take this as an identity and not concern ourselves with its rigorous justification based on generalised function or distribution theory. The variable $\tilde{x}$ is often termed an ``auxiliary field" in physics, and we shall adopt this convention here in keeping with this literature.

In certain circumstances the application above appears as part of the Hubbard-Stratonovich transform,
\begin{align}\label{eq:hubbardStratonovich}
    \exp(-\frac{a}{2} x^2) = \frac{1}{\sqrt{2 \pi a}} \int_{-\infty}^{\infty} \exp(-\frac{y^2}{2a} + ixy) dy
\end{align}
which can be viewed as an integral identity in its own right, or equivalently as a combination of the Fourier representation of the delta and the standard Gaussian integral identity. This can be derived by writing, $ \exp(-\frac{a}{2} x^2) =  \int_{-\infty}^{\infty} \exp(-\frac{a}{2} w^2) \delta(x-w)$, and subsequently using the Fourier representation \eqref{eq:fourier_dirac} and then taking the Gaussian integral over $w$. This particular application of the Dirac delta is used several times in the derivation below.

\subsection{The method of steepest descents}

The above delta trick and Hubbard-Stratonovich transformation are typically combined in physics with the method of steepest descents or saddlepoint approximation \cite{bender1999advanced}. One can view this is a powerful generalisation of Laplace's method, in that one can approximate an integral of the following form,
\begin{align}
    \int_{C} f(z) \exp(N g(z)) dz
\end{align}
where $C$ is a contour of the complex integral, and $f$ and $g$ are analytic functions of $z$. Like Laplace's method, the idea is roughly that as $N\to \infty$, the contribution to the integral is dominated by a small region about the critical point of $g$ (assuming there is a unique critical point). In the complex domain of course there are no maxima or minima and only saddlepoints. This is a powerful and now standard tool in physics, with its utility long recognised in statistics \cite{daniels1954saddlepoint}.

\subsection{Finite Dirichlet means via the Beta-Gamma algebra}

A key difference between the random neural networks and Bellman equation under uncertainty is that we consider sums of Dirichlet random variables, that is, the uncertain transition probabilities. In order to carry out the dynamic mean field calculation, one must average a moment generating function involving linear combinations of the transition probabilities, with respect to the Dirichlet uncertainty.

Linear combinations of Dirichlet variables, also known as finite Dirichlet means \cite{pitman2018random} \cite{hjort2005exact}, were studied in the context of time series analysis by \cite{vonNeumann1941} who presented an expansion for the moment generating function of these means. Here we will show that this expansion can be derived by use of the Beta-Gamma algebra property \cite{ConniffeSpencer}, \cite{pitman2018random}. 

Consider Dirichlet variables $\{P_i\}_{i=1, \dots, N}$ with aritrary parameters $\mathbf{\alpha} = \{\alpha_1, \dots, \alpha_N \}$. Given a  vector $x\in \mathbb{R}^N$, denote the finite Dirichlet mean as $M = \sum_{i=1}^N x_i P_i$. The Beta-Gamma algebra property can be stated as follows. Consider a Gamma random variable $\gamma$ with shape parameter $\alpha_0 = \sum_{i=1}^N \alpha_i$ and scale r rate parameter arbitrary, and a set of independent Gammas with parameters matching the Dirichlet parameters, that is $\gamma_i$  with $\alpha_i$. The Beta-Gamma algebra property ensures the following equality in distribution,
\begin{align}
    \gamma \sum_j x_j P_j = \sum_j x_j \gamma_j \label{eq:betaGamma}
\end{align}
That is, if you multiply a finite Dirichlet mean by an independent Gamma random variable, whose parameter matches the sum of the Dirichlet parameters, this product of two random variables is statistically identical to a linear combination of independent Gamma random variables. See \cite{pitman2018random} and \cite{ConniffeSpencer} for details on this quite remarkable result.

We can use this result to calculate the non-central moments, and in turn the central moments or cumulants directly, by taking first powers and then expectations of \eqref{eq:betaGamma}, using the assumed independence properties. See the appendix for details. We note that this approach circumvents the quite heavy derivation based on complex analysis \cite{vonNeumann1941}, which itself precludes the more standard approach of directly expanding the characteristic function, as in the Fourier inverse or saddlepoint approximation based derivations of the central limit theorem \cite{tao2012topics}. For completeness, we also mention the related central limit theorems or asymptotic normality results for functions of uniform spacings \cite{pyke1965spacings}, which are themselves distributed as flat Dirichlet variables. This line of work is largely built on Le Cam's conditional limit theorem for in this setting \cite{le_cam1958theoreme}, whose first two moments match those given by the approach of \cite{vonNeumann1941} but cannot be systematically expanded to higher moments.

\section{Dynamic mean field theory for Bayesian reinforcement learning}\label{sec:DMFT_BayesRL}

In this section we present the generating functional approach, systematically deriving a dynamic mean field theory for dynamic programming following \cite{helias2019statistical} and \cite{crisanti2018path}, using the tools described in the previous section. In the RL case, the disorder has a non-zero mean, as opposed to the spin glass and neural network cases, though see \cite{takayama1986dynamical} for a clear and careful derivation with non-zero means in the spin glass case. For the Bellman equation, we restrict our derivation to the discounted infinite horizon case. Policy evaluation and the finite horizon cases have near identical derivations. 

The derivation proceeds as follows. In order to write down a moment generating function (MGF), we first define a probability distribution over Q-value iterates by artificially injecting i.i.d Gaussian noise, of arbitrarily small variance, at each state-action site and iteration. Second, we represent the empirical distribution over iterates with a Dirac delta function, from which we then write down the moment generating function for a realisation of the disorder. By switching to the Fourier representation of the Dirac delta \eqref{eq:fourier_dirac}, it is possible to analytically calculate the average over the disorder in the system. This disorder appears as linear combinations of Dirichlet transitions, or finite Dirichlet means, whose generating function expansion we can calculate due to the Beta-Gamma algebra property \eqref{eq:betaGamma}. In the flat Dirichlet case, where the distribution over the simplex is uniform, we anticipate a saddlepoint approximation or method of steepest descents, which is possible to apply once two auxiliary fields are introduced to ``decouple" the interacting variables in the generating function. The resulting saddlepoint equations give rise to the mean field equations.

\subsection{Probability distribution over Q-value iterates}

Dynamic mean field theory considers as a starting point a moment generating function (MGF) of the distribution over trajectories of a system. Given that the system we study is deterministic, albeit subject to disorder due to uncertain parameters, it is convenient to consider a noise-perturbed system. The noise terms are needed for a large deviations analysis in the neural network case, since these provide a well defined Radon-Nikodym derivative \cite{moynot2002large}, which in continuous time is calculated by the Girsanov theorem \cite{arous1995large}. We do not have such technicalities here since the equations evolve on a discrete index, but we do need the noise to write down a probability distribution.

The perturbations are independent Gaussian noise $W_{sa}^{k}$ terms added to each state action pair $(s,a)$ and at each iteration $n$.  Thus, we study the following system 
\begin{align}\label{eq:Q_val_iter_noise}
    Q_{sa}^{k+1} = \rho_{sa} + \beta\sum_{s'} P_{s'|sa} \phi^{k}_{s'}+W_{sa}^{k}
\end{align}
 For shorthand we write the maximum over actions as $\phi^{k}_{s'}= \max_{a'} Q_{s'a'}^{k}$. This is of course the standard value function $V_{s'}^{k}$, but we maintain the notation introduced to emphasise the non-linearity and connections to the neural network literature. The distribution over a trajectory from iteration $k= 1$ to $k=M$ can be written as
\begin{align}
    &p(Q^{1:K} ) = \int  D\mathbf{W} \prod_{(sa),k}p(W^{k}_{sa}) \delta(Q^{k+1}_{sa} \! -\! \rho_{sa}-\beta\sum_{s'} P_{s'|sa}\phi^{k}_{s'} - W^{k}_{sa} ) \label{eq:discrete_path_prob_defn_Qval}
\end{align}
where we have introduced the shorthand notation $\int D\mathbf{W}  := \prod_{k=1}^{K-1} \prod_{sa} \int dW^{k}_{sa}$. If we take this average we obtain a product over Gaussian densities, which is the discrete time analogue of the Onsager-Machlup path integral \cite{helias2019statistical}. However in order to average over the disorder we introduce a response field $\tilde{Q}^{k}$ by representing the Dirac delta by its Fourier integral \eqref{eq:fourier_dirac}. Thus we have the discrete time analogue of the path integral formalism of Martin-Siggia-Rose-De Dominicis-Janssen (MSRDJ),
\begin{align}
    &p(Q^{1:K}) = \int D \mathbf{\tilde{Q}}  Z_W(-\tilde{Q}^{k}_{sa}) \exp \big[\tilde{Q}^{k}_{sa}(Q^{k+1}_{sa}- \rho_{sa} -\beta\sum_{s'} P_{s'|sa}\phi^{k}_{s'} )  \big] \label{eq:discrete_prob_dist_aux_field_Qval}
\end{align}
 where $ Z_W(-\tilde{Q}^{k}):=\int dW \exp(-\tilde{Q}^{k}_{sa} W^{k}_{sa} ) p(W^{k}_{sa})$ is by definition the generating function for the noise, and we wriute again $\int D \mathbf{\tilde{Q}} = \prod_{k=1}^{K-1} \prod_{sa}\int_{-i\infty}^{i\infty} \frac{d\tilde{Q}^{k}_{sa}}{2\pi i}$ (noting that for convenience we have moved the imaginary $i$ in the Fourier representation of the delta by a simple change of variables). We remark that some authors equivalently achieve this representation via the Hubbard-Stratonovich transformation \eqref{eq:hubbardStratonovich}.

\subsection{Moment generating function and the disorder average}

We now consider the MGF for Q-value iterates by taking the expectation of terms $\exp(j^k_{sa} Q^{k}_{sa})$ for all iterations $k$ and state action paris $(sa)$,
\begin{align}
    Z[\mathbf{j}] = \prod_{(sa), n} \int dQ^{k}_{sa} \exp(\sum_{k,sa} j^k_{sa} Q^{k}_{sa}) p(Q^{1:K})
\end{align}
We can of course calculate moments of the system via functional derivatives $\frac{\delta}{\delta j^{k}} Z[\mathbf{j}]|_{j=0} = \langle Q^{k}\rangle$. Additionally, by including a small perturbation $-\tilde{j}^{k}_{sa}$ to the iteration \eqref{eq:Q_val_iter_noise}, we can probe dependencies across iterations\footnote{For example, correlations across time between state-action values, in the finite horizon MDP.} by calculating the so called response function. While not used here explicitly, we mention linear response for completeness, as in principle it also allows us to correct the mean field theory to second order, with higher order corrections leading to Feynman diagram techniques \cite{helias2019statistical}. Using the notation $\mathbf{j}^T \mathbf{Q} = \sum_{sa, n}  j^{k}_{sa} Q^{k}_{sa}$, we write
\begin{align}
    &Z[\mathbf{j}, \mathbf{\tilde{j}} ] = \int D \mathbf{\tilde{Q}}  D \mathbf{Q} Z_W(-\tilde{Q}^{k}_{sa}) \exp \bigg[\sum_{k,sa}\tilde{Q}_{sa}^{k}(Q_{sa}^{k+1}  -\rho_{sa} - \beta\sum_{s'} P_{s'|sa} \phi_{s'}^{k}) + \mathbf{j}^T \mathbf{Q} +\mathbf{\tilde{j}}^T \mathbf{\tilde{Q}} \bigg] \label{eq:Qvalue_mgf_P_main}
\end{align}
If we assume the moment generating function is self-averaging with respect to the disorder \cite{helias2019statistical}, \cite{crisanti2018path}, then from its average we can calculate the statistics of typical realisations of the system. In taking the average, the term involving $\rho_{sa}$ in \ref{eq:Qvalue_mgf_P_main} presents no difficulty, contributing only the moment generating function of $\rho_{sa}$ which we denote $Z_{\rho_{sa}}(-\tilde{Q}^{k}_{sa})$. In taking the average over the Dirichlet, we require the moment generating function of a \emph{finite Dirichlet mean},
\begin{align}
    &\int \prod_{(sa)} dP_{\cdot|sa} p(P_{\cdot|sa}) \exp \big( - \beta \! \sum_{(sa),s'} P_{s'|sa}  \sum_{k=1}^{K-1} \tilde{Q}_{sa}^{k} \phi_{s'}^{k} \big) = \prod_{(sa)} \mathbb{E}_{M_{sa}}\exp(- M_{sa} )\label{eq:avg_Dirichlet_MGF}
\end{align}
where we have defined the finite Dirichlet mean $M_{sa} =\beta\sum_{s'} P_{s'|sa}  \sum_{k=1}^{K-1} \tilde{Q}_{sa}^{k} \phi_{s'}^{k}$. We can now write out an asymptotic expansion of the generating function of a finite Dirichlet mean, originally developed by \cite{vonNeumann1941}. The mean and variance of $M_{sa}$ are given by,
\begin{align*}
&\bar{M}_{sa} =  \beta\sum_{k=1}^{K-1} \sum_{s'} \bar{P}_{s'|sa}\tilde{Q}_{sa}^{k} \phi(Q_{s':}^{k}) \\
&\text{Var}(M_{sa}) = \beta^2\sum_{s', s''}  C_{s',s''|sa}\big( \sum_{k=1}^{K-1} \tilde{Q}_{sa}^{k} \phi_{s'}^{k} \big) \big( \sum_{m=1}^{K-1} \tilde{Q}_{sa}^{k} \phi_{s''}^{k} \big)
\end{align*}
In the flat prior case, the Dirichlet mean is $\bar{P}_{s'|sa} = \frac{1}{N}$ and the covariance $C_{s',s''|sa}$ has positive diagonal elements of order $\mathcal{O}(N^{-2})$ and negative off-diagonal of order $\mathcal{O}(N^{-3})$. Thus, the variance term $\sigma^2_{S_{sa}}$ under a flat Dirichlet is of order $\mathcal{O}(N^{-1})$, meaning that interaction terms do not contribute in the large $N$ limit. From here we can write the disorder averaged generating function for Q-value iteration as
\begin{align}
    \langle{Z}[\mathbf{j},\mathbf{\tilde{j}} ] \rangle_{P_{sa}}&= \int D \mathbf{\tilde{Q}} D \mathbf{Q} Z_W(-\tilde{Q}^{k}_{sa}) \exp \big(S[\mathbf{\tilde{Q}} \mathbf{Q}]\big)
    \label{eq:disorder_avg_Qvalue_mgf_main}\\
     S[\mathbf{\tilde{Q}}, \mathbf{Q}] &= \sum_{k=1}^{K-1} \sum_{sa}\tilde{Q}_{sa}^{k}Q_{sa}^{k} + \ln Z_{\rho_{sa}}(-\tilde{Q}^{k}_{sa})- \beta\frac{1}{N} \sum_{s'} \tilde{Q}_{sa}^{k} \phi(Q_{s':}^{k})+ \mathbf{j}^T \mathbf{Q} +\mathbf{\tilde{j}}^T \mathbf{\tilde{Q}}) 
\end{align}
The term inside the exponential $S[\mathbf{\tilde{Q}},  \mathbf{Q}]$ is known as the \emph{action} in physics, albeit in the continuous time analogue of our setting \cite{graham1977path} \cite{durr1978onsager} \cite{eyink1996action}. Importantly, we see that the action does not decompose into sums of states or state-actions, meaning the moment generating function does not factorise over state-action pairs. This is due to the term $\tilde{Q}_{sa}^{k}$ multiplying the sum over $s'$. If the MGF did factorise, we would conclude immediately that all state-action values were independent.

At this point we can follow the standard dynamic mean field theory method to ``decouple" the system and obtain the factorisation of the action over all state and action indices. This implies the asymptotic independence of the state-action values. We do this by defining a new variable and enforcing it via the Dirac delta \eqref{eq:fourier_dirac}, as described in the derivation of the Hubbard-Stratonovich transformation \eqref{eq:hubbardStratonovich}.

\subsection{Auxiliary fields and saddlepoint approximation}

Define the following auxilliary field
\begin{align}
    \theta_{1}^k :=  \frac{\beta}{N A} \sum_{s'} \phi(Q_{s':}^{k})
\end{align}
we will replace this variable within the expression above via the Dirac delta, which is represented via its Fourier form with conjugate auxiliary field $\theta_2^k$,
\begin{align}
    \delta(-N A \theta_1^k + \beta\sum_{s'} \phi(Q_{s':}^{k})) = \int_{-i\infty}^{i\infty} \frac{1}{2\pi i} d\theta_2^k \exp [\theta_2^k(-NA \theta_1^k + \beta\sum_{s'} \phi(Q_{s':}^{k})) ]
\end{align}
Dropping the terms $ \mathbf{j}^T \mathbf{Q} , \mathbf{\tilde{j}}^T \mathbf{\tilde{Q}}$ and focusing on the integral over the exponential of the action, we have
\begin{align*}
&\langle \exp(S[\mathbf{\tilde{Q}}, \mathbf{Q}]) \rangle_{P_{sa}}=  \exp \big[\sum_{k} \sum_{sa}\tilde{Q}_{sa}^{k}Q_{sa}^{k} + \ln Z_{\rho_{sa}}(-\tilde{Q}^{k}_{sa})- \beta \frac{1}{N} \sum_{s'} \tilde{Q}_{sa}^{k} \phi(Q_{s':}^{k})) \big] \\
&=\int D\theta_1 \exp \big[\sum_{k} \sum_{sa}\tilde{Q}_{sa}^{k}Q_{sa}^{k} + \ln Z_{\rho_{sa}}(-\tilde{Q}^{k}_{sa})- A\tilde{Q}_{sa}^{k} \theta_{1}^k \big] \delta(-N A \theta_1^k + \beta\sum_{s'} \phi(Q_{s':}^{k}))\\
&= \int D\theta_1 D\theta_2  \exp \big[ \sum_{k} \sum_{sa}\tilde{Q}_{sa}^{k}Q_{sa}^{k+1} + \ln Z_{\rho_{sa}}(-\tilde{Q}^{k}_{sa})- A\tilde{Q}_{sa}^{k} \theta_{1}^k- NA\theta_2^k \theta_1^k + \beta \theta_2^k \sum_{s'}\phi(Q_{s':}^{k})  )\big] 
\end{align*}
with $\int D\theta_1 D\theta_2 = \prod_{k=1}^{K-1} \int_{-i\infty}^{i\infty} \frac{1}{2\pi i} d\theta_1^k d\theta_2^k$. If we now bring back in the integral over $ \mathbf{Q},  \mathbf{\tilde{Q}}$ and the other terms in \eqref{eq:disorder_avg_Qvalue_mgf_main}, we can the disorder average MGF finally as
\begin{align}
\langle{Z}[\mathbf{j},\mathbf{\tilde{j}} ] \rangle_{P_{sa}}&= \int D\theta_1 D\theta_2  \exp \big[-N A \sum_{k} \theta_2^k \theta_1^k + N A \sum_k \ln Z[\theta_1^k, \theta_2^k]  \big] \\
&=  \int D\theta_1 D\theta_2  \exp \big[N\Gamma(\theta_1, \theta_2)]  \big]\label{eq:final_mgf_effectiveAction}
\end{align}
The term $\Gamma(\theta_1, \theta_2)$ is known as the \emph{effective} action and is the starting point of any effective field theory which includes the mean field theory developed here \cite{stapmanns2020self}. At the risk of losing the reader to yet another term, we point out that the effective action itself involves a newly defined moment generating function $Z[\theta_1^k, \theta_2^k]$, which depends on the auxiliary fields. This is defined by
\begin{align}
  &(Z[\theta_1^k, \theta_2^k])^{NA} \nonumber \\
  &=  \int  D \mathbf{\tilde{Q}} D \mathbf{Q}  Z_W(-\tilde{Q}^{k}_{sa}) \exp\big[\sum_{sa}\tilde{Q}_{sa}^{k}Q_{sa}^{k+1} + \ln Z_{\rho_{sa}}(-\tilde{Q}^{k}_{sa})- A\tilde{Q}_{sa}^{k} \theta_{1}^k +\beta \theta_2^k \sum_{s'}\phi(Q_{s':}^{k}) \big] \label{eq:new_MGF_thetas}\\
  &= \prod_{s,a} \int  D \mathbf{\tilde{Q}} D \mathbf{Q}  Z_W(-\tilde{Q}^{k}_{sa}) \exp\big[\tilde{Q}_{sa}^{k}Q_{sa}^{k+1} + \ln Z_{\rho_{sa}}(-\tilde{Q}^{k}_{sa})- A\tilde{Q}_{sa}^{k} \theta_{1}^k- \beta \theta_2^k \frac{1}{A}\phi(Q_{s':}^{k}) \big]
\end{align}
The factor of $N A$ that comes out of this generating function is due to the summation over states, which have been decoupled via the auxilliary fields. A difference to the neural network and spin glasses and the Bellman equation is that there is also a factor of $A = |\mathcal{A}|$, the action space size. This is typically much smaller than $N$ and thus does not alter the method of steepest descents which we now apply.

\subsection{Saddlepoint approximation}

The leading order approximation in the method of steepest descents, or saddlepoint approximation, is to approximate the integral by the saddlepoint itself, ignoring the contributions from higher order terms (eg. the Hessian and higher terms).  We find the critical point of the effective action in equation \eqref{eq:final_mgf_effectiveAction},
\begin{align}
    \frac{\delta}{\delta \theta_{1,2}^k} \Gamma[\theta_1, \theta_2] = 0
\end{align}
We now calculate the derivatives,
\begin{align}
    \frac{\delta}{\delta \theta_{1}^k} \Gamma[\theta_1, \theta_2] &= - NA \theta_2^k + N A  \frac{\delta  \ln Z[\theta_1^k, \theta_2^k] }{\delta \theta_{1}^k} \\
    &= - NA \theta_2^k + N A  \frac{1}{Z[\theta_1^k, \theta_2^k}  \frac{\delta Z[\theta_1^k, \theta_2^k] }{\delta \theta_{1}^k}\\
    &= -NA \theta_2^k +NA \langle -A\tilde{Q}_{sa}^{k}  \rangle_{\theta_1, \theta_2}
\end{align}
where the average $\langle \cdot \rangle_{\theta_1, \theta_2}$ emerges due to the derivative of the log generating function $\ln Z[\theta_1^k, \theta_2^k] $ of equation \eqref{eq:new_MGF_thetas}. Setting this derivative to zero, we have 
\begin{align}
    \theta_2^k =  - A\langle \tilde{Q}_{sa}^{k}  \rangle_{\theta_1, \theta_2} \label{eq:theta2_SP}
\end{align}
A similar calculation reveals
\begin{align}
    \theta_1^k =   \frac{\beta}{A} \langle \phi(Q_{s':}^{k}  \rangle_{\theta_1, \theta_2}
    \label{eq:theta1_SP}
\end{align}
It turns out that at the saddlepoint $(\theta_1^*, \theta_2^*)$ the average \eqref{eq:theta2_SP} of the response field $\tilde{Q}_{sa}^{k}$ must be zero due to the normalisation of the probability distribution and that we do not fix both ends of our trajectory (only the initial point is assumed to be given).  For detailed general discussion of this quite technical point see \cite{stapmanns2020self} and in the spin glass case \cite{sompolinsky1982relaxational}.

Determining the saddlepoint $(\theta_1^*, \theta_2^*=0)$ exactly involves solving a the saddlepoint equations \emph{self-consistently}, which can roughly be thought of as something of a fixed point equation but for the statistics of the effective field theory\footnote{see section II of \cite{stapmanns2020self} for instructive examples of this notion}. In the present context, to see what this self-consistent solution looks like, let us first substitute the point  $(\theta_1^*, \theta_2^*=0)$ into \eqref{eq:final_mgf_effectiveAction}, noting the integral has now vanished since we have taken the saddlepoint approximation,
\begin{align*}
\langle{Z}[\mathbf{j},\mathbf{\tilde{j}} ] \rangle_{P_{sa}}&=  \exp \big[N\Gamma(\theta_1^*, \theta_2^*)  \big]\\
 &= \prod_{s,a} \int  D \mathbf{\tilde{Q}} D \mathbf{Q}  Z_W(-\tilde{Q}^{k}_{sa}) \exp\big[\sum_k\tilde{Q}_{sa}^{k}Q_{sa}^{k+1} + \ln Z_{\rho_{sa}}(-\tilde{Q}^{k}_{sa})- A\tilde{Q}_{sa}^{k} (\theta_{1}^k)^* \big]\\
 &= \prod_{s,a} \int  D \mathbf{\tilde{Q}} D \mathbf{Q}  Z_W(-\tilde{Q}^{k}_{sa}) \exp\big[\sum_k \tilde{Q}_{sa}^{k}Q_{sa}^{k+1} + \ln Z_{\rho_{sa}}(-\tilde{Q}^{k}_{sa})- \beta \langle \phi(Q_{s':}^{k})  \rangle_{\theta_1^*, \theta^*_2} \big]
\end{align*}
If we now consider equation \eqref{eq:discrete_path_prob_defn_Qval} and were to work backwards to equation \eqref{eq:Q_val_iter_noise}, we can see that we have an effective ``equation of motion in place of $\eqref{eq:Q_val_iter_noise}$ where the term $\beta\sum_{s'} P_{s'|sa} \phi^{k}_{s'}$ has been replaced by $\beta \langle \phi(Q_{s':}^{k})  \rangle_{\theta_1^*, \theta^*_2}$. We determine this average over $\phi(Q_{s':}^{k})$ recursively from the initial iteration and thus find our self-consistent solution. This recursive equation over iterations is our mean field equation. The following example helps to clarify things further.

\subsection{Example}

In the case that the posterior over the mean reward $\rho_{sa}$ is Gaussian, we have that its moment generating function is
\begin{align}
     \ln Z_{\rho_{sa}}(-\tilde{Q}^{k}_{sa}) = -\mu_{\rho_{sa}}\tilde{Q}^{k}_{sa} +\frac{1}{2} \sigma^2_{\rho_{sa}}(\tilde{Q}^{k}_{sa})^2
\end{align}
substituting this expression in the averaged MGF, as evaluated at the saddlepoint we find
\begin{align*}
\langle{Z}[\mathbf{j},\mathbf{\tilde{j}} ] \rangle_{P_{sa}}&= \prod_{sa} \int  D \mathbf{\tilde{Q}} D \mathbf{Q}  Z_W(-\tilde{Q}^{k}_{sa}) \exp\big[\sum_k \tilde{Q}_{sa}^{k}Q_{sa}^{k+1} - \mu_{\rho_{sa}}\tilde{Q}^{k}_{sa} +\frac{\sigma^2_{\rho_{sa}}}{2} (\tilde{Q}^{k}_{sa})^2- \beta \langle \phi(Q_{s':}^{k})  \rangle \big]
\end{align*}
which we can identify as the statistics corresponding to a Gaussian process over the state-action values. Explicitly, the mean is given by 
\begin{align*}  
\mathbb{E} \big(Q^{t+1}_{s,a}\big) & = \mu_{\rho_{sa}} + \beta \mathbb{E}\big( \max_{a'} Q^k_{s'a'}\big).
\end{align*}
where the expectation of the maximum with respect to $ Q^k_{s'a'}$, whose mean is given by the previous iteration, and whose variance is just that of the mean-reward, $\sigma^2_{\rho_{sa}}$. The result thus takes a deceptively simple form, and is similar to the neural network and spin glass cases, however there is no correlation across time of the Gaussian process, with only the mean propagating forward across iterations. In general of course, the process will not be Gaussian since the mean-rewards can have arbitrary distribution.

\section{Dynamic mean field programming}\label{sec:DMFP}

The mean field equation, which we call dynamic mean field programming (DMFP), for Q-value iteration is given by
\begin{align}
    \mu_{sa}^{k+1} = \mu_{\rho_{sa}}
    + \beta \mathbb{E}_{Q^{k}}\max_{a'} Q^{k}_{s'a'}
    \label{eqn:DMFP_valIter}
\end{align}
where we denote the mean of $ Q^{k+1}_{sa}$ as $\mu_{sa}^{k+1}$, and the expectation $\mathbb{E}_{Q^{k}}$ is with respect to the distribution of the mean rewards $\rho_{s'a'}$ with mean shifted by $\mu_{sa}^{k}$. The presence of the maximum requires us to turn to approximations. Here we turn to extreme value theory and bounds on the expected maximum, which we discuss in the next section.

In the case of policy evaluation, there is no need for approximations. Given a randomized policy $\pi_{a|s}$, the policy evaluation equations for the value function are given by
\begin{align}\label{eq:policy_eval}
    V_{s}^{\pi, n+1} = \sum_a \pi_{a|s} \big( \rho_{sa} + \beta \sum_{s'} P_{s'|sa} \  V_{s'}^{\pi,n} \big)
\end{align}
Denoting the posterior mean for the value functions  $\mu_{s}^{k} = \mathbb{E} V_{s'}^{\pi,n}$, as for the Q-values, the policy evalutation DMFP equations are given by
\begin{align}
    \mu_{s}^{\pi,n+1} = \sum_a \pi_{a|s} \big(  \mu_{\rho_{sa}}
    + \beta \sum_{s'} \bar{P}_{s'|s,a} \  \mu_{s'}^{\pi,n+1} \big)
    \label{eqn:DMFP_polEval}
\end{align}


\subsection{Approximate statistics of DMFP}
In order to compute the solutions to the optimal Q-values via DMFP, one needs to calculate the first moment of the maxima over the actions of $Q^{k}_{sa}$, which in general are non-identical and of arbitrary distribution, with the distribution being that of the mean-rewards, shifted by the propagated means. As a first approximation, we appeal to extreme value theory, which describes the distribution of the maximum over asymptotically many variables (appropriately re-scaled). This theory was originally developed for i.i.d variables, but there exist generalisations to non-independent variables in the case the mean-rewards are Gaussian \cite{extreme_value_falk2010laws}, \cite{husler1994extremes}. This work establishes that in the Gaussian case the distribution approaches a Poisson limit, from which we may calculate the desired moments.

In the i.i.d case extreme value theory provides closed form approximate equations. Assuming the Dirichlet parameters are set to $\alpha_{s'|sa} = 1$, then if we use a Type-I extreme value or Gumbel distribution as an approximation, we have
\begin{align}
    \mu^{k+1} &= \mu_\rho +\beta \big(\mu^{k} +\sigma_\rho(b_{|\mathcal{A}|}+a_{|\mathcal{A}|} \gamma_{\text{EM}}) \big) \label{eq:evt_mean}
\end{align}
where $\gamma_{\text{EM}}$ is the Euler-Mascheroni constant, and $a_{|A|}$ and $b_{|A|}$ are constants dependent on the action space size $|A|$, given in \cite{fisher_tippett_1928}. The approximation can be improved by using a Type-III or Weibull distribution \cite{fisher_tippett_1928} or other approximations \cite{Cohen82_penultimate}.

If we wish to weaken the assumption that the mean-reward and transition dynamics priors are identically distributed, we may appeal to bounds on the expected maximum. We have considered the relatively simple bounds of \cite{aven1985upper} and the tight bounds of \cite{bertsimas2006tight}, as well as a simple lower bound via Jensen's inequality, $\mathbb{E} \max_i X_i \geq \max_i \ m_i$, for random variables $X_i$ with arbitrary dependence and means $m_i$. A simple heuristic which we also consider, indeed which inspired the paper initially, is based on algorithms which iterate modified Bellman equations that include an ``exploration bonus" for each state-action pair. For random variables $X_i$ with mean $m_i$ and standard deviation $\sigma_i$, we write the approximation as $\mathbb{E} \max_i X_i \approx \max_i \ m_i +\sigma_i$.  We see that this is bounded below by Jensen's inequality and above by Aven's bound. We view this as a heuristic approximation of the expected maximum over the Q-values. Other such approximations are possible.

\begin{figure}    
    \centering
    \includegraphics[scale=0.45]{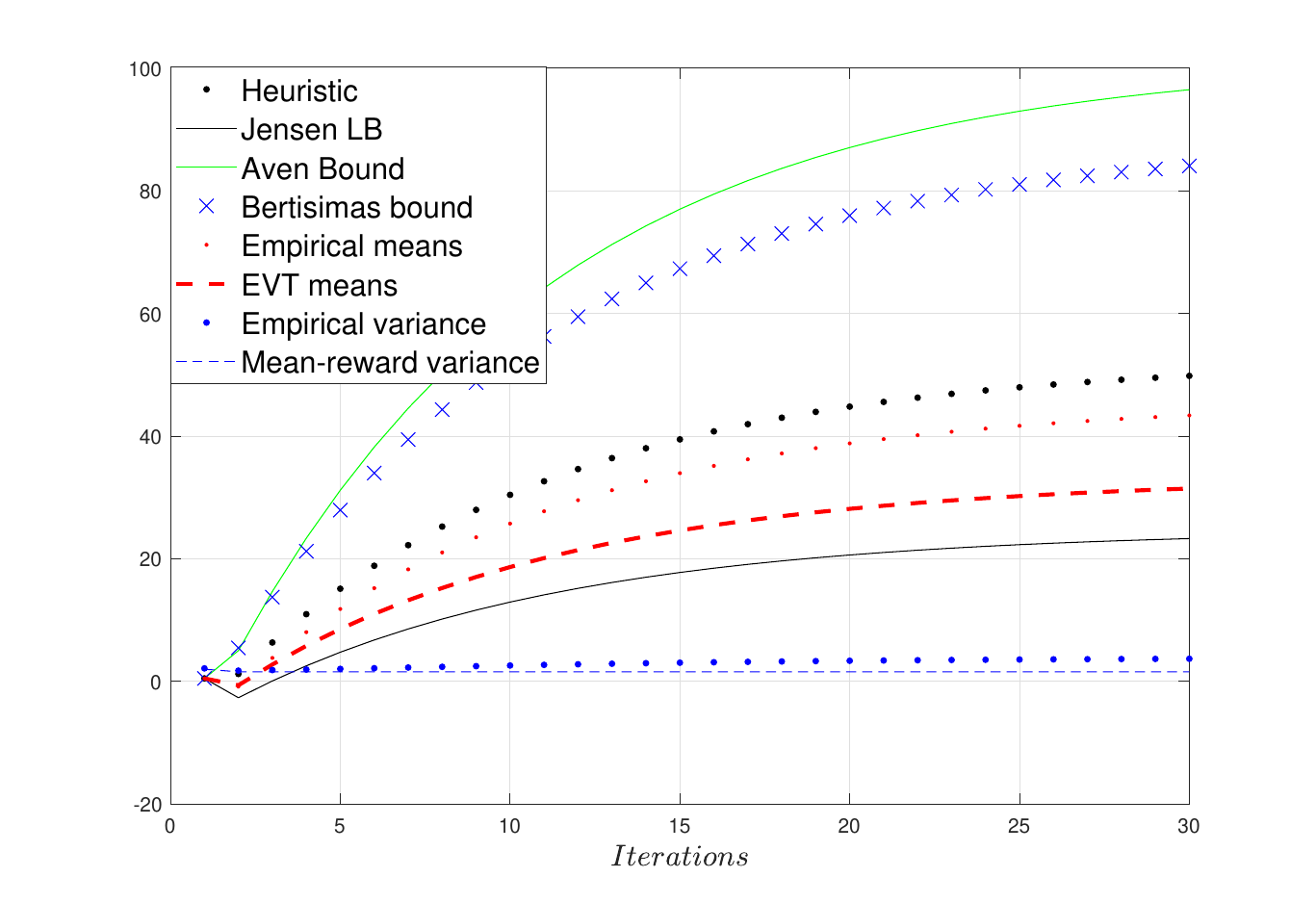}
    \label{fig:EVT_bounds_DMFP_noniid}
    \caption{DMFP simulations in the non-identical mean-reward case, for a discount factor of $\beta=0.9$, $k=250$, $|\mathcal{A}|=10$, $k=300$ samples. Overlaid are the extreme value theory (EVT), bounds and heuristic approximations. The empirical mean is given by red dots, and the empirical variance by blue dots. We see the empirical variance is small but not vanishing, owing to fluctuations from finite size effects.}
\end{figure}

The predictions of the DMFP theory and accuracy of the approximations are shown in Figure \ref{fig:EVT_bounds_DMFP_noniid}, where strong agreement is observed. The simulations are straightforward; we draw from given Dirichlet and mean-reward distributions, just as in Thomspon sampling, and run value iteration until convergence. The simulations presented are for a small system with only $|\mathcal{S}|=50$ and $|\mathcal{A}|=30$, with $\alpha_{sa} = 1$ over the statespace, and mean-rewards with identical uniform priors. we present the performance of the bounds, EVT approximation and the heuristic, in the non-iid case, for one particular Q-value.

\section{Discussion}\label{sec:discussion}

The dynamic mean field theory developed in this paper provides precise, analytic insight into the properties of Markov decision processes, in terms of how statistical assumption on the MDP parameters translate to the value functions. In this section we discuss the interpretations and implications of the results and the breakdown of the mean field theory, the next possible technical challenges, as well as existing related work in both reinforcement learning and neural network theory literature.

\subsection{ Interpretations of the theory}

A headline result of the paper is that the distribution over Q-values factorises across state-action pairs in the asymptotic limit of number of states, under the assumptions of Section \ref{sec:bayesian_RL}. This is a surprising result, since the recurrent structure clearly introduces dependencies between Q-values. Of course, there is a dependency on the level of the mean field statistics, but given these statistics, as computed by the DMFP equations, the Q-values are statistically independent.

Let us first reflect on assumptions 1 and 2, the statistical independence of the parameters across state-action pairs, and the flat Dirichlet assumptions, respectively. It is common in the literature to have the independence assumption in the model based approaches to tabular MDPs. Likewise, the flat Dirichlet is a common assumption to have satisfied at the start of learning, as a prior. Of course, this will not hold throughout learning for MDPs with sparse transition matrices.

This means that these algorithms start by assuming all Q-values are independent, which would appear to be a strong assumption to start learning with\footnote{We expect a similar independence structure to exist in the frequentist setting, where we consider \emph{estimators} for the optimal Q-values based on point estimates for the MDP parameters.}. The dependence between parameters at different state-actions should be contrasted to the linear quadratic regulator, the cornerstone of continuous state control theory. In the linear quadratic case, the state dynamics and cost functions are typically not state-dependent functions. This reflects a causal model of the real world. For example, we expect the dynamics of a vehicle to be quite similar in different locations, despite differences in terrain.

As discussed in the introduction, the mean field theory result provides practical insight even at the start of learning. If we make these prior assumptions, this is equivalent to assuming an agent faces a set of $N$ independent multi-armed bandit problems in each state. If a practitioner is happy with these assumptions, then this should be exploited, with an agent acting locally and not planning ahead, since the mean field theory may approximately hold for some time around learning. An alternative of course is to make different prior assumptions should be changed.

\subsection{Breakdown of the theory}

How might the mean field theory break down? Here we speculate that some of the most relevant ways are if the MDP and the belief structure has (i) finite state space size $N$, (ii) sparsity in transition functions, (iii) statistical dependence between state-action parameters (mean-rewards and transition dynamics), or (iv) unbounded horizon (eg. scaling with $N$). 

In practice we will have finite size effects (i) of some form. If the independence assumptions are maintained, we may study the finite size effects perturbatively \cite{stapmanns2020self} \cite{segadlo2022unified}. A theoretical description of this breakdown may give further insight into the phenomenology of reinforcement learning problems and the dependence between value functions.

Sparsity in the MDP transition matrices (ii) is natural property to consider. To an extent this might be viewed as a finite size effect, and captured by that theory, however other techniques may be relevant in this setting. In paritcular, certain limits exist for interacting particle models on sparse \emph{random} graphs that are relevant \cite{maclaurin2020large} \cite{oliveira2020interacting} \cite{lacker2023local}.

Relatedly, we can report that the assumption of improper priors for the Dirichlet transitions leads to a breakdown in the theory, resembling more closely the Gaussian process behaviour of neural networks \cite{crisanti2018path} \cite{segadlo2022unified}. Specifically, if we scale the Dirichlet parameters for each transition probability as $\alpha_{s'|sa} = \frac{1}{N}$ then the variance for the finite Dirichlet mean is non-vanishing. This sort of scaling anticipates a high degree of sparsity in the transitions. Of course, the validity of the above saddlepoint approximation may be called into question.

The introduction of statistical dependence between parameters (iii) means that the average in equation \eqref{eq:avg_Dirichlet_MGF} cannot be taken for each state-action pair. One would expect more dependence between the state-action value functions and a loss of the self-averaging property (technically, this would be a loss of the large deviations principle for the empirical measure of trajectories). In this case the problem may be closer to the original spin glass cases \cite{crisanti1987dynamicsSpherical} \cite{crisanti1988dynamicsIsing} rather than more recent neural network theory. In the spin glass case the couplings have a specific dependence, namely they are symmetric: the coupling $J_{ij}$ between spin $i$ and $j$ is equal to the reverse coupling $J_{ji}$. Of course, with all the couplings being positive, it is possible that the phenomenon of frustration that exists in spin glasses will not appear (in physicists language, positive couplings are to ferromagnetic). Thus the loss of self-averaging may not be accompanied by glassy behaviour in the dynamics.

The length of the horizon (iv), or effective horizon in the infinite discounted MDP, may bring on different regimes of behaviour as well. In the neural network literature the scaling of horizon (or rather, depth) with width simultaneously has been associated with heavy tailed phenomena \cite{hanin2020products}, which we can report within experiments in this paper.

Finally, we conclude by posing the natural question of what type of MDP satisfies the flat Dirichlet assumption throughout learning, rather than just the start. We can conclude that such a system is a simple contextual bandit with a discrete, one dimensional context; the state arrives i.i.d, with no Markov dynamics present. In this case, the reinforcement learning problem corresponds to a set of $N$ independent multi-armed bandit problems.

\subsection{Further technical challenges}

The asymptotic independence result for the Bellman equation may be surprising to many in reinforcement learning or statistical physics. A rigorous proof of the DMFT result, such as those based on large deviations analysis \cite{arous1995large} and \cite{moynot2002large}, is needed. A next step from the field theoretic perspective is to consider finite size corrections to the mean field theory, as well as the introduction of sparsity to test the theory. We expect this to require first an auxiliary field and subsequently a saddlepoint approximation \cite{helias2019statistical}. From the saddlepoint approximation one may then calculate next-to-leading-order corrections \cite{segadlo2022unified}, and further diagrammatic or reinformalisation group based corrections \cite{stapmanns2020self} \cite{grosvenor2022edge}.



As a final comment, we anticipate the theory presented has implications for frequentist as well as Bayesian approaches. In Bayesian statistics, one considers a random variable of interest, for which a prior is assigned and posterior distributions calculated given data. From this posterior one can produce various estimates. In frequentist statistics, one can also produce estimates from \emph{estimators}, which have fluctuations. A good example of the difference is the Kalman filter, which can be considered a Bayesian estimator for the state of a linear quadratic Gauss-Markov system, or a minimum variance estimator from the frequentist perspective. Visiting the problem of designing better estimators for the Q-value functions would be an interesting line of work. The theory may help in the construction of tighter confidence intervals for UCB-like approaches \cite{OsbandVanRoyPSRLvsOptimism}.


\section*{Acknowledgements}
The author would like to acknowledge the support of Prof. Hung Nguyen, DST Group and the many helpful discussions from students and colleagues, Blake Donnelly, Lachlan MacDonald, Ian Fuss, Kelli Francis-Staite, Langford White, Blake Bordelon, Federica Gerace, Carlo Lucibello and Jack Valmadre.

\newpage
\appendix
\onecolumn

\section{Moment generating function expansion of linear combination of Dirichlet}

We can write down the moments of linear combinations of Dirichlet variables,  $M = \sum_i P_i x_i$ using a fundamental and remarkable property of the so called Beta-Gamma algebra \cite{pitman2018random}, 
\begin{align}
    \gamma \sum_i P_i x_i = \sum_i \gamma_i x_i
\end{align}
where the Dirichlet variables $\{ P_i \}$ are $\{\alpha_i\}$ and $\gamma$ is a Gamma random variable independent of the $P_i$. with parameter $\alpha = \sum_i$ and likewise $\gamma_i$ has parameter $\alpha_i$.

We can write the expansion as follows using $\log(1+x) \approx x - x^2/2 + x^3 /3 + ...$
\begin{align}
    \langle \exp( tM) \rangle_M  &= 1+ t\Ebb_M M + \frac{1}{2}t^2 \Ebb_M M^2 + \frac{1}{3 \!}t^3 \Ebb_M M^3 +... \\
    &= \exp( \log(1+ t\Ebb_M M + \frac{1}{2}t^2 \Ebb_M M^2 + \frac{1}{3 \!}t^3 \Ebb_M M^3 +... ))\\
    &\approx \exp( t\Ebb_M M + \frac{1}{2}t^2 \Ebb_M M^2 + \frac{1}{3 \!}t^3 \Ebb_M M^3 + t^2\Ebb_M M + \frac{1}{2}t^2 \Ebb_M M^2 + \frac{1}{3 \!}t^3 \Ebb_M M^3))\\
    &=  t\Ebb_M M + \frac{1}{2}t^2 \Ebb_M M^2 + \frac{1}{3 !}t^3 \Ebb_M M^3 - \frac{1}{2} \bigg( t\Ebb_M M + \frac{1}{2}t^2 \Ebb_M M^2 + \frac{1}{3 !}t^3 \Ebb_M M^3\bigg)^2
\end{align}
But this is nothing more than the cumulant generating function, for which the first three terms are the first three central moments. Now writing $\Ebb_M M = \ol{M}$, $\Ebb_M M^2 = \ol{M^2}$, 
\begin{align}
    \exp( \log \langle \exp( tM) \rangle_M ) &\approx  t\ol{M} + \frac{1}{2}t^2 (\ol{M^2}- \ol{M})^2 + \frac{1}{3 ! }t^3 \ol{(M - \ol{M})^3} 
\end{align}

If The Dirichlet parameters are $\alpha_i = 1$, then we have
\begin{align}
    \ol{M} &= \frac{1}{N} \sum_i x_i\\
    \ol{M^2} &= \frac{2}{N(N+1)} \sum_{i,j} x_i x_j \\
    \ol{M^3} &= \frac{6}{N(N+1)(N+2)} \sum_{k} x_k^3 +\frac{6}{N(N+1)(N+2)} \sum_{i\neq j} x_i x_j^2 +\frac{1}{N(N+1)(N+2)} \sum_{i\neq j\neq k} x_i x_j x_k =
\end{align}
The variance is given by
\begin{align}
    \ol{ M^2} -\ol{M}^2 &= \frac{N-1}{N^2(N+1)} \sum_{i} x^2_i - \frac{1}{N^2(N+1)} \sum_{i\neq j} x_i x_j \\
\end{align}

\section{ Simulations.}

\begin{figure}[ht]  
    \centering
    \includegraphics[scale=0.42]{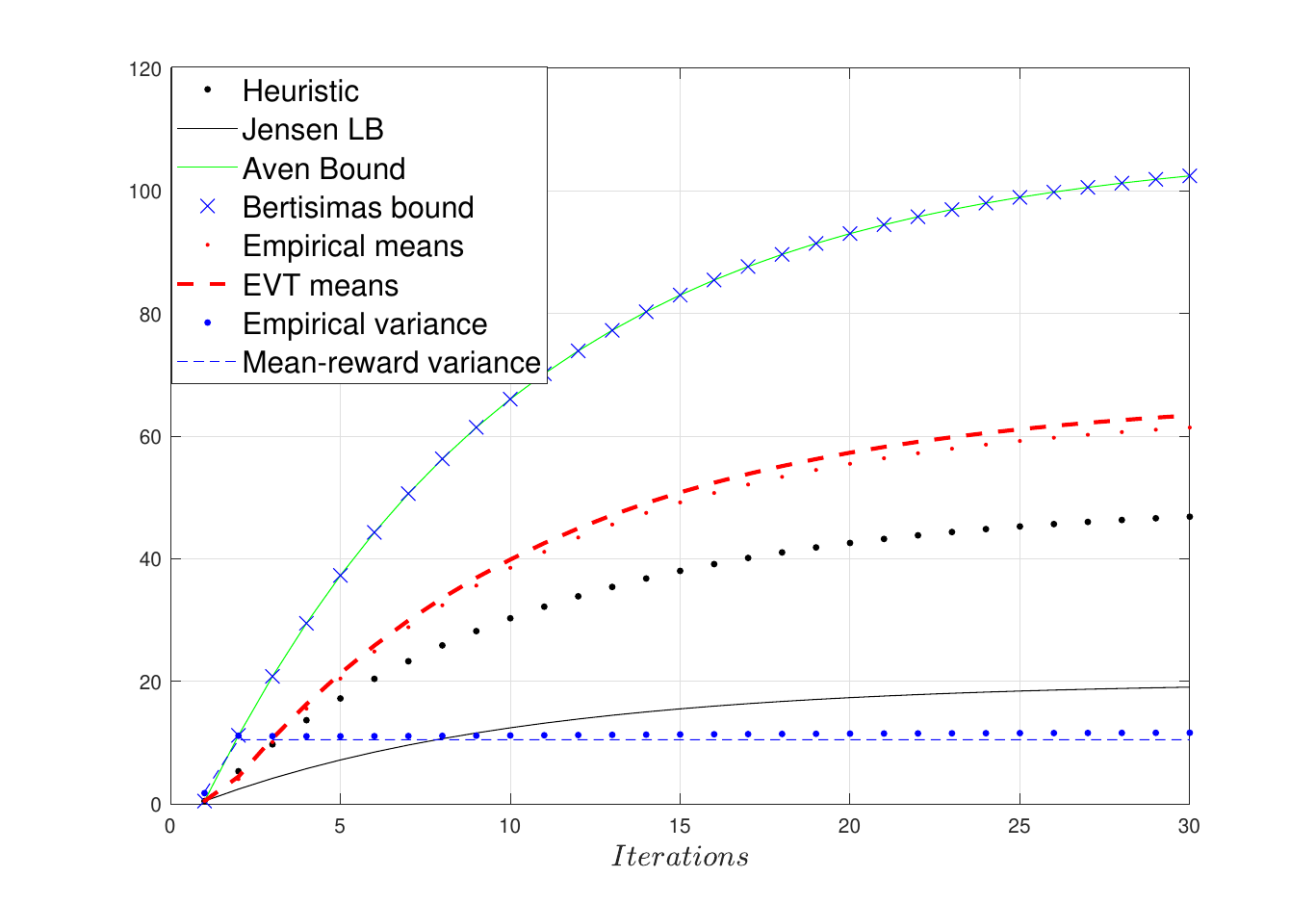}
        \includegraphics[scale=0.42]{nonidentical_plots-_1_.pdf}
    \caption{DMFP simulations of Q-value iteration in the i.i.d case (Top) and non-identical case (Bottom), for a discount factor of $\beta=0.9$, $k=250$, $|\Ac|=10$, $k=300$ samples. Overlaid are the extreme value theory (EVT), bounds and heuristic approximations. In the identical case we see the Aven and Bertsimas et al. bounds agree, while the EVT approximation will become exact as the action space grows. The heuristic inspired by exploration bonuses is seen to remain quite accurate compared to the bounds across both cases, for the systems considered here.}
   \label{fig:DMFP_Qval_bound_approx}
\end{figure}

We present further simulation evidence in this appendix of the validity of the DMFP equations and the underlying theory. In figure \ref{fig:DMFP_Qval_bound_approx} we present the performance of the bounds and approximations of the DMFP equations, as well as the empirical estimates of the optimal Q-value iterates, in the i.i.d mean-reward case. The second figure below is the non-identical case presented in the main paper, repeated for ease of comparison. We see the extreme value approximation is very accurate here.

\begin{figure}[ht]
    \centering
    \includegraphics[scale=0.7]{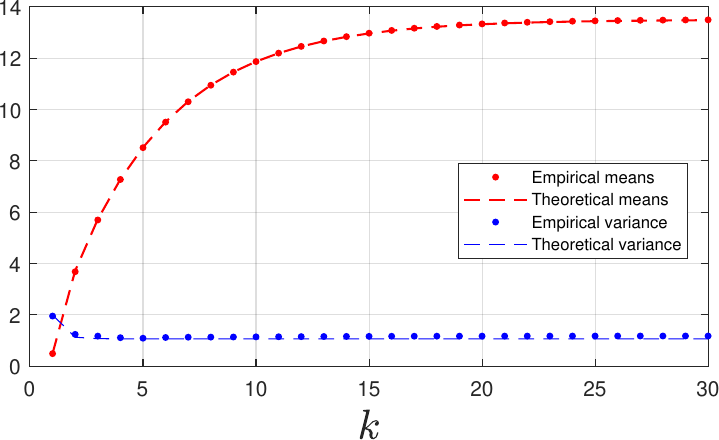}
        \includegraphics[scale=0.69]{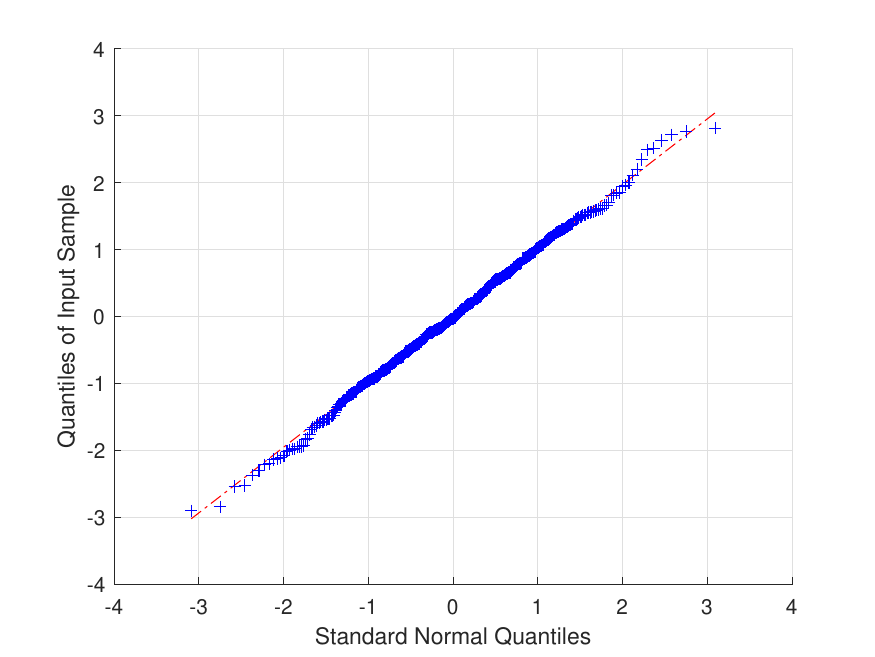}
    \caption{DMFP simulations of Q-value iteration in the i.i.d case, for an MDP of size $|\Sc| = 50$, $|A|=20$ with identical flat Dirichlet priors for the transitions for each state action pair, with identical Gaussian priors for the mean-rewards, and discount factor $\beta=0.9$.  Left: Mean and variance for a Q-value $Q_{sa}$, with the empirical mean and variance given by dots, extreme value theory approximations given by dashed lines. Right: Q-Q plots for the term $\sum_{s'} P_{s'|sa} \max_{a'} Q_{s'a'}$, against normal quantiles.}        \label{fig:DMFP_theoryEmpirical_QQplot}
\end{figure}

In figure \ref{fig:DMFP_theoryEmpirical_QQplot} we present simulations in the i.i.d case with a Gaussian prior for the mean-rewards. The first figure again shows the accuracy of the extreme value approximation, the second presents a QQ-plot that demonstrates the accuracy of the theory; the distribution of the Q-values is that of the immediate mean-reward with a shifted mean. In this case, this distribution is Gaussian.

\pagebreak
\section{Extreme value theory approximations}\label{sec:appendixB_EVT}

 Due to the discontinuous and nonlinear function $\max(\cdot)$ in the value iteration schemes, the distributions of maxima of Q-values must be studied through extreme value statistics \cite{fisher_tippett_1928,Cohen82_penultimate}. This allows analytical expressions for the mean and variance of the Q-values to be derived. Using a type I extreme value distribution for Gaussian random variables allows the moments of the maxima to be calculated \cite{fisher_tippett_1928,Cohen82_penultimate}. 
 
 \noindent
 Defining the mean as $\mu(k)=\mathbb{E}\left[ Q^k(s,a)\right]$ and variance as $\nu(k)=\text{Var}(Q^k(s,a))$, the mean field iteration functions are 
\begin{align}
    \mu(k+1)&=\bar{r}+\beta \left(\mu(k)+\left(b_{|\mathcal{A}|}+a_{|\mathcal{A}|}\gamma_{\text{EM}}\right)\sqrt{\nu(k)}\right) \label{eqn:mu}\\
    \nu(k+1)&=\sigma^2_r+\beta^2\frac{\pi^2}{12}a_{|\mathcal{A}|}\nu(k)\quad, \label{eqn:nu}
\end{align}

\noindent
where $\gamma_{\text{EM}}$ is the Euler-Mascheroni constant, and $a_{|\mathcal{A}|}$ and $b_{|\mathcal{A}|}$ are positive constants dependent on the action space size $|\mathcal{A}|$ given by solving the expressions
\begin{align}
    &b_{|\mathcal{A}|}\sqrt{2\pi} \exp (0.5(b_{|\mathcal{A}|})^2) = |\mathcal{A}| \label{eqn:b} \\ 
    &a_{|\mathcal{A}|}=\frac{b_{|\mathcal{A}|}}{(b_{|\mathcal{A}|})^2+1}\quad. \label{eqn:a}
\end{align}

These constants increase as the size of the action space increases. The accuracy of the extreme value distribution can be improved through using a type III Weibull distribution \cite{fisher_tippett_1928}.

\subsection{Extreme value theory for non-identical case}

In the case that we consider non-identically distributed Dirichlet transition probabilities $P_{sa}$ and non-identical mean rewards $\rho_{sa}$, we thus have that the Q-values will be independent but non-identical Gaussian variables. This poses a challenge for calculating the statistics of the maximum $\max_a Q_{sa}$.

An asymptotic approximation would appeal to extreme value theory, however this theory was derived in the case where one takes $n$ \emph{samples} of from a population \cite{fisher_tippett_1928}. While closed form approximations exist in the case of non-identical means and identical variances, we would of course like to consider both non-identical means and variances. For this, we must resort to numerical approximations. Consider a sequence of Gaussian random variables 
\begin{align*}
    \{X_i\}_{i \geq 1}
\end{align*}
with non-constant mean values \(\mathbb{E}(X_i) = \mu_i\) and variances \(\Var(X_i) = \sigma_{i}^2\). Now suppose that we have another sequence of Gaussian random variables: \(\{Y_i\}_{i\geq 1}\), which is stationary. We can then define from this a non-stationary Gaussian sequence as:
\begin{align*}
    \{X_i\}_{i\geq 1} = \{\mu_i + \sigma_i Y_i\}_{i\geq 1},
\end{align*}
from which the maximum, \(M_n = \sup{X_i \mid i \leq n}\) gives us the relationship
\begin{align*}
    \{M_n \leq u\} = \left\{Y_i \leq \frac{u-\mu_i}{\sigma_i} \mid i \leq n \right\}
\end{align*}
where \(u\) is a bound for a stationary Gaussian sequence. The CDF is given by:
\begin{align*}
     P (M_n \leq u) = \prod_{i=1}^{k}P(X_i\leq u) = \prod_{i=1}^{k}P(Y_i\leq u_i) = \prod_{i=1}^{k} \Phi (u_{i})
\end{align*}
due to the independence. From the lemma below we know that there exists a real value \(\tau\), depending on $\{(\mu_i, \sigma_i^2) \}_i$ such that this product tends to the function \(e^{-\tau}\). 
that there exists a real value \(\tau\), such that this product tends to the function \(e^{-\tau}\). 

\begin{lemma}\citet{extreme_value_falk2010laws}
Let \(\{u_{ni}, 1 \leq i \leq n, n = 1, 2, ...\}\) be constants such that \(\lambda_n = \min_{1\leq i\leq n} u_{ni} \to \infty\). Then for \(0 \leq \tau \leq \infty\),
$\prod_{i=1}^{k} \Phi (u_{ni}) \to e^{-\tau(u,\mu_i, \sigma_i) }, \quad \text{ as } n \to \infty$
if and only if $\sum_{i=1}^{k} (1 - \Phi(u_{i})) \to \tau(u,\{\mu\}_i, \{\sigma\}_i) \quad \text{ as } n \to \infty$.
\end{lemma}


We have the CDF of the maximum of non-identical Gaussians, which are independent (or possibly weakly dependent). From the CDF, we can calculate the mean and variance of the maximum. This can be done by standard methods (ie. Tonelli's theorem).

\bibliographystyle{abbrvnat}
\bibliography{bibliography}

\end{document}